\documentclass[10pt, a4paper]{article}

\usepackage[]{lrec-coling2024} 
\usepackage{amsmath}
\title{Identifying Narrative Patterns and Outliers in Holocaust Testimonies Using Topic Modeling}

\name{Maxim Ifergan$^\dagger$\, Renana Keydar$^\ddagger$ , Omri Abend$^\dagger$\, Amit Pinchevski$^\diamondsuit$} 

\address{$^\dagger$ Department of Computer Science $^\ddagger$ Faculty of Law and Digital Humanities \\ $^\diamondsuit$ Department of Communication and Journalism \\ Hebrew University of Jerusalem\\
         \{{first\_name\}.\{last\_name\}@mail.huji.ac.il}}

\abstract{
The vast collection of Holocaust survivor testimonies presents invaluable historical insights but poses challenges for manual analysis. This paper leverages advanced Natural Language Processing (NLP) techniques to explore the USC Shoah Foundation Holocaust testimony corpus. By treating testimonies as structured question-and-answer sections, we apply topic modeling to identify key themes. We experiment with BERTopic, which leverages recent advances in language modeling technology. We align testimony sections into fixed parts, revealing the evolution of topics across the corpus of testimonies. This highlights both a common narrative schema and divergences between subgroups based on age and gender. We introduce a novel method to identify testimonies within groups that exhibit atypical topic distributions resembling those of other groups. This study offers unique insights into the complex narratives of Holocaust survivors, demonstrating the power of NLP to illuminate historical discourse and identify potential deviations in survivor experiences.
\\ \newline  }

\begin{document}

\maketitleabstract

\section{Introduction}

In recent decades, significant efforts have been made to gather the accounts of the remaining Holocaust survivors. The passing of the last living witnesses and the beginning of the era of the post-testimony occurs simultaneously with technological developments in NLP.The wealth of testimonies in the archives presents a challenge: how to preserve the significance of individual stories within a vast collection of a thousand testimonies, while also giving voice to the collective body of testimonies in a manner that honors the individuality of each story. By employing techniques such as contextualized topic modeling and topic narrative analysis, we aim to uncover broad trends within the collection, while ensuring the preservation of the uniqueness and integrity of each personal narrative.

Despite advancements in NLP, representation of long texts still poses a challenge to state-of-the-art models \citep{piper-etal-2021-narrative, castricato-etal-2021-towards, mikhalkova-etal-2020-modelling, Dong2023ASO}. \citet{Antoniak2019NarrativePA} pioneered the representation and visualization of narratives as sequences of interpretable topics. And while previous topic modeling analyses of Holocaust testimonies \citep{Blanke2019UnderstandingMO} have provided valuable insights, they treated the corpus as a monolithic body of text, obscuring the unique narrative structure of individual testimonies. Furthermore, using non-contextualized topic modeling such as LDA (\citep{Blei2001LatentDA}) treated the text as a body of words without order. Recent advancements in topic modeling techniques such as BERTopic \citep{Grootendorst2022BERTopicNT}, and other Contextualized Topic Modeling \citep{Bianchi2020PretrainingIA, Angelov2020Top2VecDR, Pham2023TopicGPTAP} leverage language model representation to better identify and predict the text topics. While such methods were applied to Holocaust testimonies \citep{Wagner2022TopicalSO}, the main focus was on the segmentation of the testimonies for topic modeling. 
Our contributions are as follows:
\begin{itemize}
\item
    We apply a novel contextualized topic modeling approach, BERTtopic, to holocaust testimonies, revealing the main themes and their distribution.
\item
    We examine the evolution of topics across aligned sections of testimonies, revealing a typical narrative scheme.
\item
    We investigate how age and gender are expressed in the narrative structure of testimonies, highlighting distinctions between survivor subgroups.
\item
    We introduce a novel method for identifying divergent testimonies, i.e., testimonies within a given group that exhibit atypical topic distributions, resembling patterns more characteristic of other groups. We demonstrate it in a case-study of different age groups.
\end{itemize}

We note that related contributions appear in an unpublished paper of ours (under review; anonymized) that uses an earlier contextualized topic model \citep[CTM;][]{Bianchi2020PretrainingIA} for a similar process. The current paper uses a better performing model \citep{Grootendorst2022BERTopicNT} regarding topic diversity and coherence and a more detailed and precise narrative analysis approach.

\section{Corpus Level Statistics}

\begin{figure}[t]
\centering
\includegraphics[scale=0.25]{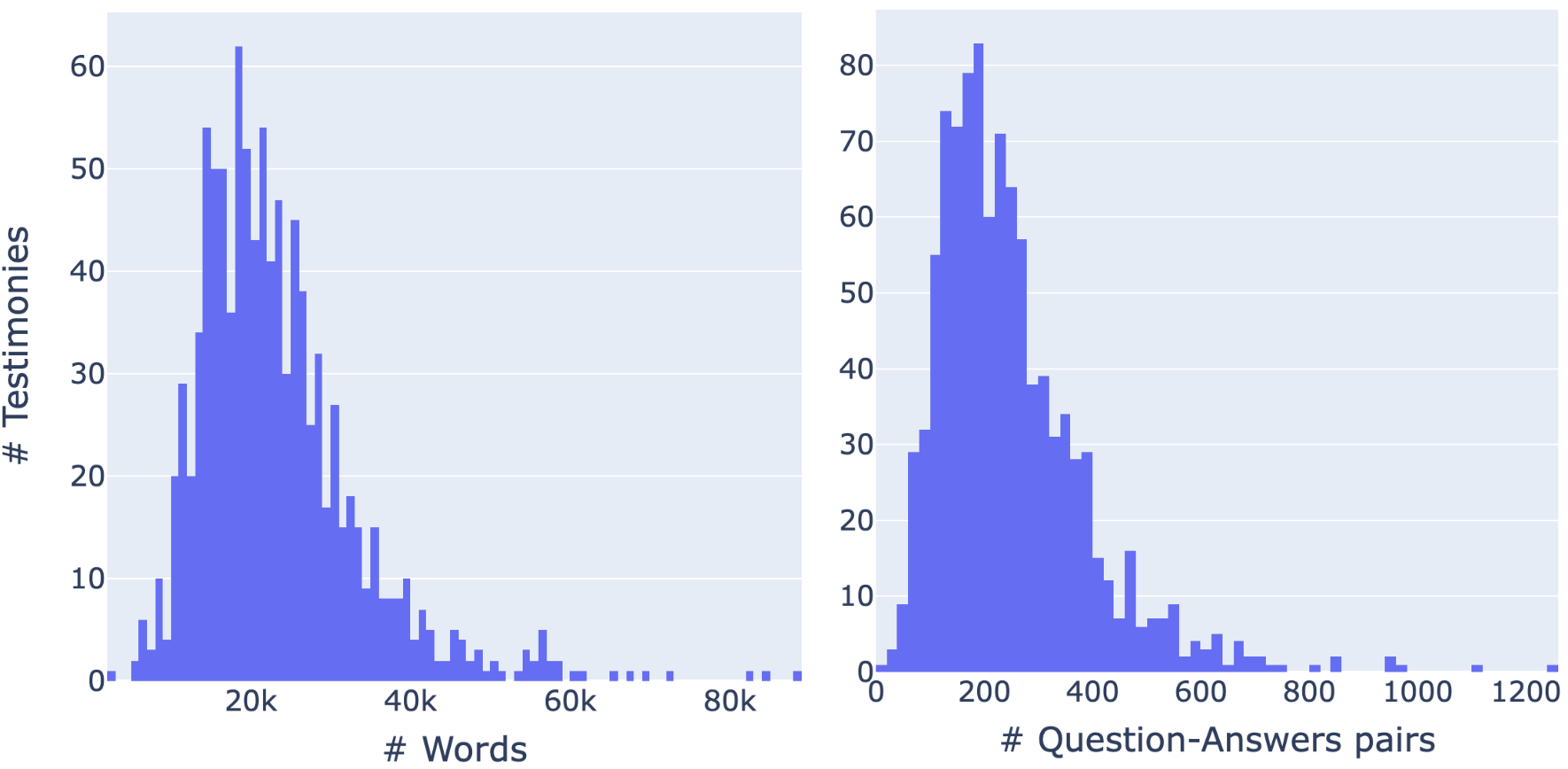}
\caption{Testimonies number of words and number QA-pairs histogram.} \label{fig: figure1}
\end{figure}

\begin{figure*}[t]
\centering
\includegraphics[scale=0.34]{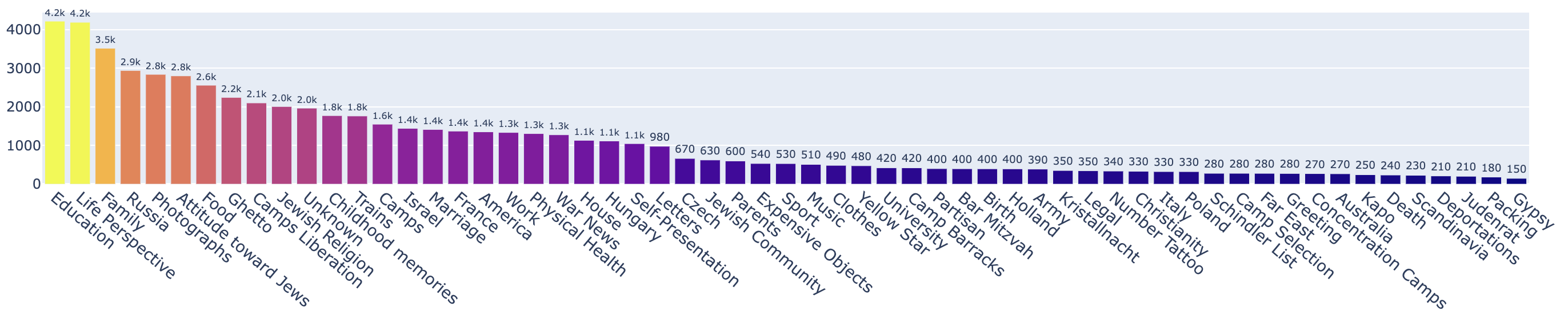}
\caption{Corpus level QA-s topics histogram.} \label{fig: figure2}
\end{figure*}

This paper analyzes transcripts from the USC Shoah Foundation, a corpus containing 1000 oral testimonies in English. Survivors originated from over 30 countries, with a significant representation from Poland and Germany. The testimonies were recorded between 1996 and 2015, offering insights into the survivors' experiences decades after the events of the Holocaust. The length of the testimonies ranges from 3K to 88K words, with a mean length of 23K words. Each testimony contains an average of 250 questions, with the majority of question-answer pairs (95$\%$) consisting of no more than 400 words. Fig. \ref{fig: figure1} illustrates the distribution of testimony lengths.

\begin{figure*}[t]
\centering
\includegraphics[scale=0.315]{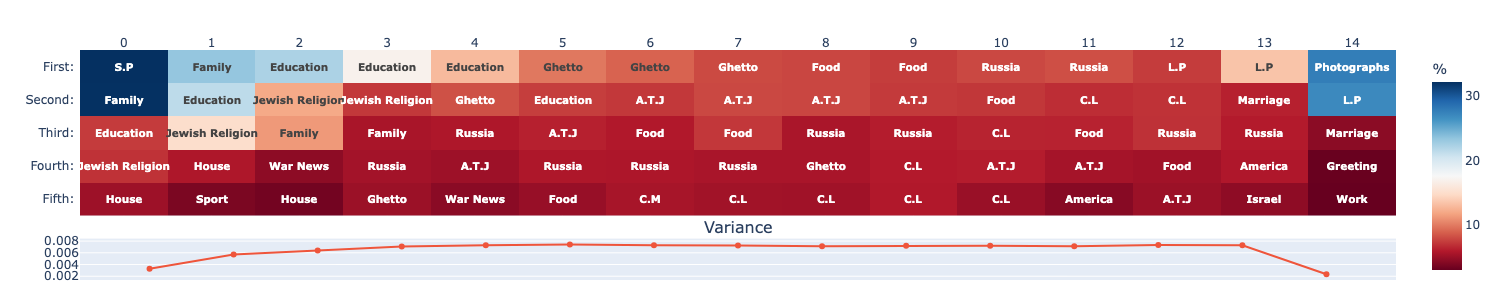}
\caption{The 5 most prevalent topics and topics variance for each part. A.T.J = Attitude toward Jews, S.P = Self-Presentation, C.L = Camps Liberation, L.P = Life Perspective, and C.M = Childhood Memories} \label{fig: figure3}
\end{figure*}

\section{BERTopic: Topic Analysis}

We use BERTopic to identify the topics within the corpus. Preprocessing involves the merging of consecutive very short sections (question-answer pairs <200 words) and the division of very long sections (>450 words) to mitigate potential outlier effects. 
BERTopic leverages all-MiniLM-L6-v2 \citep{Wang2020MiniLMDS} document embeddings and a TF-IDF based clustering approach, providing a context-aware analysis that surpasses traditional methods like LDA \citep{Blei2001LatentDA}. For dimensionality reduction, UMAP \citep{McInnes2018UMAPUM} is employed before clustering with HDBSCAN \citep{McInnes2017hdbscanHD}. Unlike LDA, BERTopic dynamically determines the number of topics only by determining the minimum cluster size for HDBSCAN, resulting in greater flexibility. Our dataset yielded 58 topics, with approximately 4$\%$ outliers classified as ``unknown topic''. We set the minimum cluster size to be 50 sections.

To ensure interpretability, BERTopic extracts c-TF-IDF\footnote{https://maartengr.github.io/BERTopic/api/ctfidf.html} word representations from each section's cluster, revealing the importance of words within each topic. 
The most representative word is selected for initial topic representation. A domain expert then manually reviews these word sets and assigns a descriptive title to each topic, ensuring both accuracy and clarity. Notably, the topics detected by the model align with those outlined in the USC Shoah Foundation's interviewer guidelines \footnote{https://sfi.usc.edu/content/interviewer-guidelines} but also extend way beyond them. The guidelines encourage interviewers to ask about pre-war life, family, religion, politics, community, and experiences of antisemitism. The model's successful detection of these themes confirms the effectiveness in identifying core key topics.

\section{Narrative analysis}

This study analyzes individual survivor testimonies as narratives --  sequences of interpretable topics \citep{Antoniak2019NarrativePA}. We aim to construct comprehensive narratives from the corpus testimonies that enable comparisons without sacrificing their temporal structures. Several challenges arise in this analysis. First, each testimony comprises a large number of sections (250 on average), conflicting with the direct interpretation and visualization purposes. Secondly, variations in testimony length complicate direct comparisons of narrative structures.

To address this, we divide testimonies into a fixed manageable number of parts, defining the part's theme representation as the distribution of its sections' topics. This division requires considering the trade-off between preserving temporal detail and achieving clear visualization and comparison. A large number of parts yields more nuanced narratives but risks excessive details and redundancy, whereas fewer parts allow better interpretation and visualization at the expense of obscuring finer temporal shifts in topics. After careful examination and consultation with domain experts in Holocaust studies and digital humanities, we divide testimonies into 15 equal parts. This strikes a balance between the need for detail and the goals of clear visualization and comparisons of topic distribution across parts.

\begin{figure*}[t]
\centering
\includegraphics[scale=0.35]{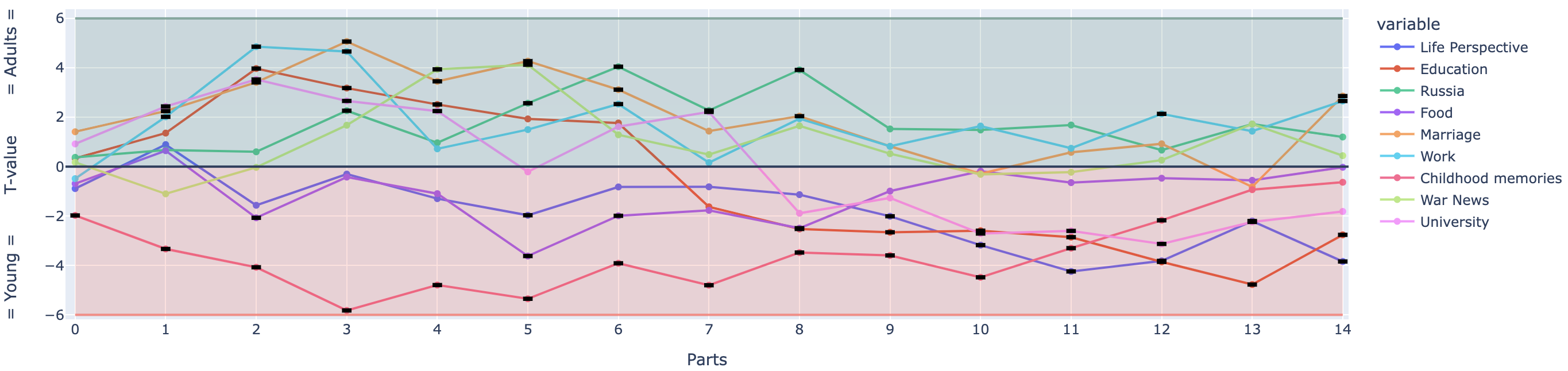}
\caption{Adults vs. young survivors typical testimony t-test. The Black Point represents values with p-values under 0.01.} \label{fig: figure4}
\end{figure*}

\subsection{Typical Testimony Narrative Schema}

This section examines the most common topics covered in each part of a Holocaust survivor testimony, as well as the variation in topic representation between the different testimony parts. The analysis is based on Fig. \ref{fig: figure3}, which shows the distribution of topics across the 15 parts into which each testimony was divided.

The first part of all testimonies is dominated by the topic of self-presentation followed by the family topic. It is perhaps unsurprising that many testimonies begin this way, as survivors introduce themselves and their families to the interviewer. The fact that the self-presentation topic rarely appears later may not constitute a significant finding, but it is nevertheless of importance as it validates the model's analytic capability. The next two parts of the testimonies also reveal a number of common topics, associated with the description of community life before the war. These include family, education, religion, house, and sport. The latter part also contains the topic of war news, hinting at the events to come. 

In contrast to the dominance of common topics at the beginning of the testimonies, the middle parts show greater variance in the topic distributions. Each part typically features several common topics with similar percentages (around 5-15$\%$). This might reflect the diversity of experiences among Holocaust survivors. The middle part topics vary starting with ghettos and war news to concentration/death camps and food, resolving in the rise in dominance of the camps liberation topic.

In the final parts of the testimonies narratives once again the model identifies a few dominant topics. These include interview-related topics such as presenting family pictures and discussing life after the Holocaust topics. Additionally, topics related to life after the war emerged, such as immigration and establishing work and marriage in new countries.

In conclusion, the BERTopic model successfully identifies a typical structure for Holocaust survivor testimonies, particularly at the beginning and end. The middle sections show more variation, reflecting the different experiences of individual survivors.

\begin{figure*}[t]
\centering
\includegraphics[scale=0.35]{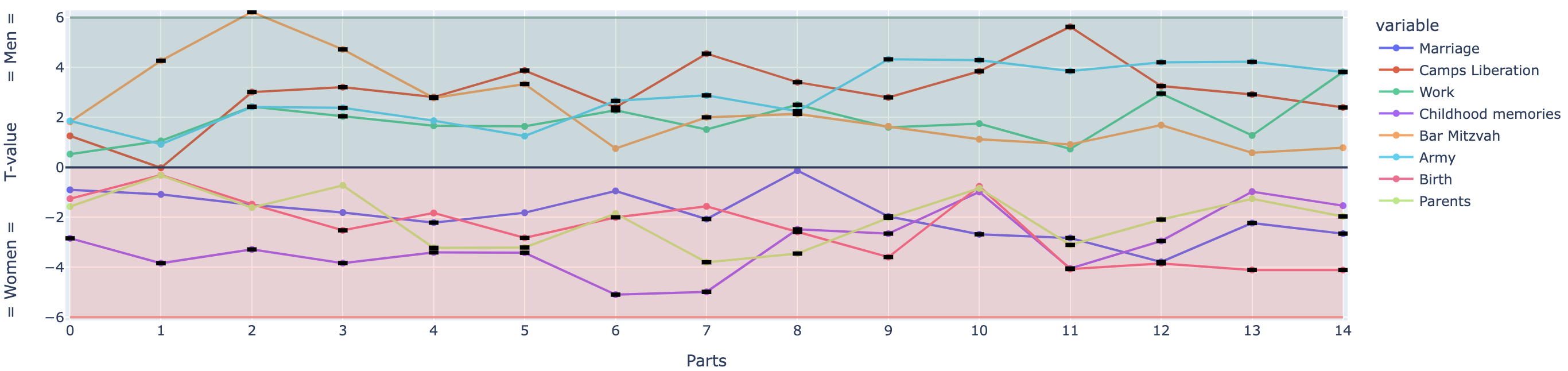}
\caption{Men vs. women survivors typical testimony t-test. The Black Point represents values with p-values under 0.01.} \label{fig: figure5}
\end{figure*}

\subsection{Gender and Age as Expressed in Testimonies Narratives}

This section introduces a method for comparing the narrative trajectories present within different groups of Holocaust survivor testimonies. We apply this method to investigate gender- and age-based differences in testimonial narratives.

To begin, we compute a typical testimony path for each group under consideration (e.g., male vs. female, young vs. older survivors). This 'typical' testimony schema represents the average topic distribution across the 15 fixed parts. Next, we perform t-tests for each part to quantify the differences in topic prevalence between groups. Topics with a substantial t-value (above 3.5$\%$) and a low probability of such deviation arising by chance (p-value under 0.01) are flagged as characteristic of the group in which they are more prevalent.

Let's consider the age-based comparison between younger survivors, born 1925-1940, experiencing the Holocaust as children (522 testimonies), and older survivors, born 1902-1925, adults during the Holocaust (467 testimonies). Fig. \ref{fig: figure4} reveals interesting distinctions. Topics like "Childhood Memories" and "Food" dominate the middle parts of younger survivors' testimonies, while  "Life Perspective" features in the final parts. Conversely, "Marriage", "Work", and "War News" are more prominent in the middle of older survivors' accounts. Interestingly, while education-related topics seem more prevalent at the beginning of older survivors' testimonies, they tend to re-emerge near the end for the younger group. Turning to the gender-based analysis, with a balanced corpus of 531 male and 469 female testimonies, Fig. \ref{fig: figure5} highlights potential differences. Topics like "Bar Mitzvah", "Army", "Camp Liberation", and "Work" are more characteristic of men's testimonies. In contrast, "Birth", "Childhood Memories", "Parents", and "Marriage" are more prevalent in women's testimonies.  This analysis reveals how men and women may structure their narratives differently, particularly in the middle sections of their testimonies.

The USC Shoah Foundation's interviewer guidelines do not provide specific instructions for ordering topics or tailoring questions based on the subject's age or gender. This suggests that the observed differences in narrative structure between these groups are not a direct result of the guidelines. Rather, they may stem from the interviewers' individual approaches or the survivors' unique experiences and perspectives.

\section{Exploratory Study Identifying Diverging Narratives}

\begin{figure*}[t]
\centering
\includegraphics[scale=0.38]{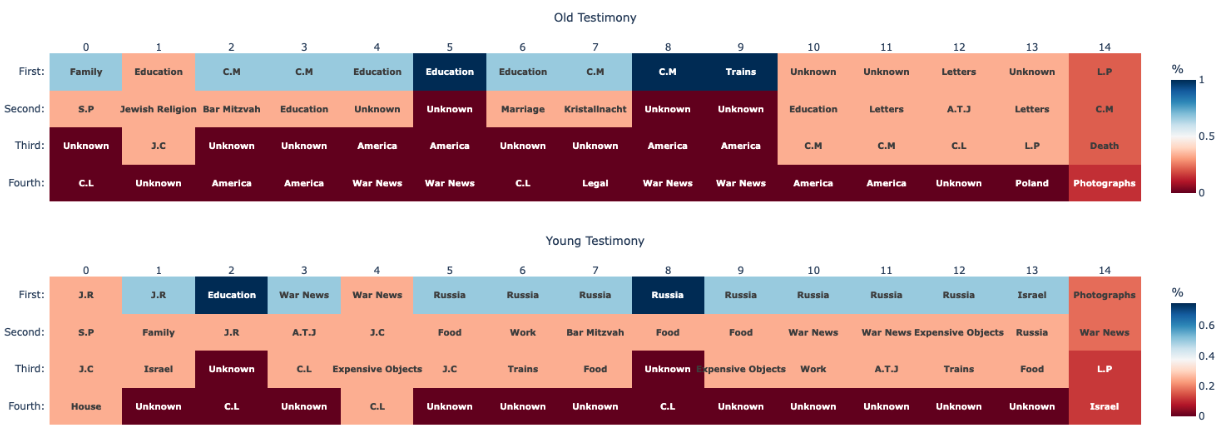}
\caption{Outliers testimonies. A.T.J = Attitude toward Jews, S.P = Self-Presentation, C.L = Camps Liberation, L.P = Life Perspective, J.C = Jewish Community, J.R = Jewish Religion, and C.M = Childhood Memories
} \label{fig: figure7}
\end{figure*}

This study introduces a novel method for identifying testimonies within a specific group that exhibit topic distribution patterns more characteristic of another group. Our goal is to pinpoint narratives that stand out as atypical within their designated category. 
We achieve this by defining a scoring function that quantifies the similarity between a testimony's topic distribution and the typical narrative patterns of a different group.
Let us formalize the scoring function which yields a high score for testimonies from A that resemble the narrative patterns typical of B. 
Let $t = (t_{1}, t_{2}, ..., t_{15})$ represent the testimony's topic distributions from group A, where each $t_{i}$ is a vector of topic probabilities for part $i$. And, $C_A = \{(i_1, j_1), (i_2, j_2), ..., (i_n, j_n)\}$ denotes the characteristic topic-part pairs for group A, where $i_x$ is a part index and $j_x$ is a topic index. 
These pairs have high t-values (>3.5) and low p-values (<0.01) in the group comparison. $C_B$  similarly represents the characteristic topic-part pairs for group B.

$$R_{B}=\sum_{(i,j)\in C_B} t_{i}[j] \cdot |Tvalue_{B}(i,j)|$$
$$R_{A}=\sum_{(i,j)\in C_A} t_{i}[j] \cdot |Tvalue_{A}(i,j)|$$
$$S(t, C_A, C_B)= R_{B} - R_{A} $$

Finally, we apply an argmax operation to spot those testimonies exhibiting the highest resemblance to group B's typical narrative:

$$\text{argmax}_{t\in A}  S(t, C_A, C_B)$$

\begin{figure}[t]
\centering
\includegraphics[scale=0.32]{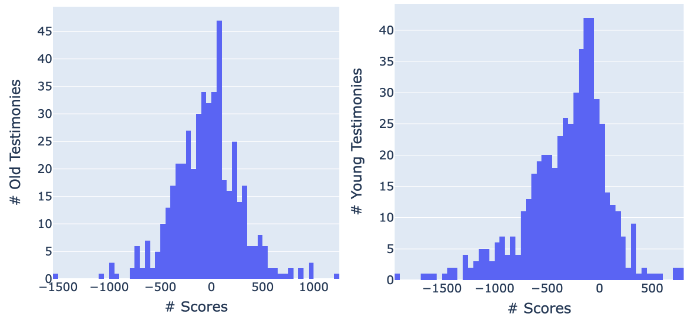}
\caption{Testimonies scores histogram.} \label{fig: figure6}
\end{figure}

When comparing older and younger survivor groups, Fig. \ref{fig: figure6} presents the distribution of resemblance scores for testimonies within the groups. The uneven distribution favoring negative scores reveals that higher scores tend to be related to non-conforming narratives. Using this method,  Fig. \ref{fig: figure7} highlights two specific examples: a younger survivor whose narrative strongly resembles the older group, and vice-versa emphasizing topics characteristic of the opposite group.

\section{Conclusion and Future Work}
This study applies NLP techniques to explore the complex narratives within the USC Shoah Foundation's Holocaust testimonies. Contextualized topic modeling with BERTopic reveals key themes and their distributions within the corpus. And, by aligning testimonies into fixed parts, we unveiled a common narrative trajectory along with age- and gender-based variations. Our method detects divergent testimonial narratives, identifying those within one group that exhibit topic patterns characteristic of another group. Future Work will extend the analysis by comparing survivor narratives across corpora\footnote{Yale Fortunoff Archive} and other testimonial archives to identify both shared and distinct narratives patterns.

\section{Acknowledgements}
The authors acknowledge the USC Shoah Foundation - The Institute for Visual History and Education for its support of this research. This research was supported by grants from the Israeli Ministry of Science and Technology and the Council for Higher Education and the Alfred Landecker Foundation.

\section{Bibliographical References}\label{sec:reference}

\bibliographystyle{lrec-coling2024-natbib.bst}
\bibliography{lrec-coling2024-example.bib}

\clearpage

\appendix
\section{Topic list}\label{appendix: A}
\vspace{.2cm}
\centering
\begin{tabular}{ll}
\multicolumn{1}{c}{\textbf{Topic Title}} & \multicolumn{1}{c}{\textbf{Top-15 words in topic}} \\ \hline
\multicolumn{1}{|l|}{Life Perspective} & \multicolumn{1}{l|}{\begin{tabular}[c]{@{}l@{}}children', 'think', 'holocaust', 'thank', 'say', 'life', 'grandchildren', \\'years', 'want', 'experiences', 'people', 'much', 'god', 'thats', 'never'\end{tabular}} \\ \hline
\multicolumn{1}{|l|}{Photographs} & \multicolumn{1}{l|}{\begin{tabular}[c]{@{}l@{}}picture', 'taken', 'thank', 'photograph', 'left', 'name', 'right', 'photo', \\'next', 'daughter', 'son', 'crew', 'sister', 'year', 'wife'\end{tabular}} \\ \hline
\multicolumn{1}{|l|}{Family} & \multicolumn{1}{l|}{\begin{tabular}[c]{@{}l@{}}name', 'mother', 'father', 'born', 'family', 'mothers', 'remember', 'sister', \\'brother', 'fathers', 'lived', 'years', 'sisters', 'brothers', 'grandfather'\end{tabular}} \\ \hline
\multicolumn{1}{|l|}{Education} & \multicolumn{1}{l|}{\begin{tabular}[c]{@{}l@{}}school', 'jewish', 'antisemitism', 'hitler', 'jews', 'remember', 'teacher', \\'hebrew', 'yiddish', 'german', 'friends', 'language', 'teachers', 'polish', 'years'\end{tabular}} \\ \hline
\multicolumn{1}{|l|}{Russia} & \multicolumn{1}{l|}{\begin{tabular}[c]{@{}l@{}}russian', 'russians', 'russia', 'people', 'away', 'germans', 'army', 'take', \\'train', 'told', 'come', 'want', 'food', 'says', 'already'\end{tabular}} \\ \hline
\multicolumn{1}{|l|}{Jewish Religion} & \multicolumn{1}{l|}{\begin{tabular}[c]{@{}l@{}}synagogue', 'holidays', 'used', 'shabbos', 'remember', 'religious', \\'shabbat', 'shul', 'friday', 'father', 'passover', 'home', 'holiday', 'jewish', 'family'\end{tabular}} \\ \hline
\multicolumn{1}{|l|}{Self-Presentation} & \multicolumn{1}{l|}{\begin{tabular}[c]{@{}l@{}}name', 'born', 'birth', 'spell', 'interview', 'please', 'date', \\'english', 'today', '1997', 'conducting', 'interviewer', 'language', '1998', 'maiden'\end{tabular}} \\ \hline
\multicolumn{1}{|l|}{Ghetto} & \multicolumn{1}{l|}{\begin{tabular}[c]{@{}l@{}}ghetto', 'people', 'germans', 'jews', 'work', 'food', 'place', 'little', \\'lived', 'remember', 'jewish', 'house', 'already', 'street', 'away'\end{tabular}} \\ \hline
\multicolumn{1}{|l|}{Food} & \multicolumn{1}{l|}{\begin{tabular}[c]{@{}l@{}}food', 'bread', 'soup', 'eat', 'potatoes', 'little', 'day', 'used', 'people', \\'water', 'piece', 'work', 'something', 'hungry', 'put'\end{tabular}} \\ \hline
\multicolumn{1}{|l|}{Marriage} & \multicolumn{1}{l|}{\begin{tabular}[c]{@{}l@{}}married', 'wedding', 'met', 'husband', 'wife', 'marry', 'israel', 'years', \\'meet', 'name', 'marriage', 'laughs', 'come', 'want', 'family'\end{tabular}} \\ \hline
\multicolumn{1}{|l|}{Israel} & \multicolumn{1}{l|}{\begin{tabular}[c]{@{}l@{}}'israel', 'palestine', 'zionist', 'british', 'kibbutz', 'organization', 'jewish', \\'people', 'army', 'hebrew', 'tel', 'war', 'come', 'also', 'years'\end{tabular}} \\ \hline
\multicolumn{1}{|l|}{France} & \multicolumn{1}{l|}{\begin{tabular}[c]{@{}l@{}}'french', 'france', 'paris', 'belgium', 'people', 'brussels', 'germans', 'train', \\'german', 'antwerp', 'war', 'border', 'think', 'vichy', 'little'\end{tabular}} \\ \hline
\multicolumn{1}{|l|}{Physical Health} & \multicolumn{1}{l|}{\begin{tabular}[c]{@{}l@{}}'hospital', 'sick', 'doctor', 'typhus', 'camp', 'fever', 'people', 'doctors', \\'typhoid', 'day', 'couldnt', 'food', 'take', 'put', 'work'\end{tabular}} \\ \hline
\multicolumn{1}{|l|}{Hungary} & \multicolumn{1}{l|}{\begin{tabular}[c]{@{}l@{}}'hungarian', 'hungary', 'hungarians', 'budapest', 'jews', 'romania', 'romanian', \\'jewish', 'people', 'war','many', 'labor', 'started', 'germans', 'father'\end{tabular}} \\ \hline
\multicolumn{1}{|l|}{America} & \multicolumn{1}{l|}{\begin{tabular}[c]{@{}l@{}}'ship', 'states', 'united', 'york', 'boat', 'new', 'visa', 'america', 'come', 'arrived', \\'american', 'quota', 'affidavit', 'days', 'papers'\end{tabular}} \\ \hline
\multicolumn{1}{|l|}{Camps} & \multicolumn{1}{l|}{\begin{tabular}[c]{@{}l@{}}'auschwitz', 'birkenau', 'camp', 'people', 'gas', 'work', 'saw', 'day', 'barracks', \\'train', 'see', 'barrack', 'right', 'knew', 'women'\end{tabular}} \\ \hline
\multicolumn{1}{|l|}{House} & \multicolumn{1}{l|}{\begin{tabular}[c]{@{}l@{}}'room', 'house', 'apartment', 'kitchen', 'rooms', 'lived', 'bedroom', 'big', 'living', \\'floor', 'remember', 'home', 'dining', 'describe', 'used'\end{tabular}} \\ \hline
\multicolumn{1}{|l|}{Camps Liberation} & \multicolumn{1}{l|}{\begin{tabular}[c]{@{}l@{}}'camp', 'german', 'american', 'germans', 'saw', 'americans', 'day', 'see', \\'people', 'soldiers','planes', 'war', 'prisoners', 'started', 'liberated'\end{tabular}} \\ \hline
\multicolumn{1}{|l|}{Trains} & \multicolumn{1}{l|}{\begin{tabular}[c]{@{}l@{}}'train', 'people', 'cattle', 'trains', 'wagon', 'camp', 'march', 'long', 'water', \\'many', 'station', 'car', 'food', 'days', 'see'\end{tabular}} \\ \hline
\multicolumn{1}{|l|}{Work} & \multicolumn{1}{l|}{\begin{tabular}[c]{@{}l@{}}'job', 'business', 'worked', 'work', 'money', 'working', 'company', 'store', 'bought', \\'make', 'years', 'week', 'started', 'factory', 'want'\end{tabular}} \\ \hline
\multicolumn{1}{|l|}{Attitude Toward Jews } & \multicolumn{1}{l|}{\begin{tabular}[c]{@{}l@{}}'says', 'jewish', 'told', 'come', 'jews', 'mother', 'want', 'away', \\'german', 'see', 'knew', 'people', 'look', 'gave', 'man'\end{tabular}} \\ \hline
\multicolumn{1}{|l|}{Music} & \multicolumn{1}{l|}{\begin{tabular}[c]{@{}l@{}}'music', 'sing', 'singing', 'song', 'songs', 'opera', 'played', 'piano', 'sang', \\'play', 'remember', 'used', 'yiddish', 'voice', 'laughs'\end{tabular}} \\ \hline
\multicolumn{1}{|l|}{Childhood memories} & \multicolumn{1}{l|}{\begin{tabular}[c]{@{}l@{}}'mother', 'father', 'remember', 'think', 'parents', 'see', 'sister', \\'never', 'really', 'knew', 'always', 'happened', 'mean', 'couldnt', 'tell'\end{tabular}} \\ \hline
\multicolumn{1}{|l|}{Czech} & \multicolumn{1}{l|}{\begin{tabular}[c]{@{}l@{}}'prague', 'czech', 'czechoslovakia', 'theresienstadt', 'train', 'people', 'see', 'come', \\'army', 'left', 'knew', 'stayed', 'transport', 'home', 'want'\end{tabular}} \\ \hline
\multicolumn{1}{|l|}{Letters} & \multicolumn{1}{l|}{\begin{tabular}[c]{@{}l@{}}'letters', 'letter', 'wrote', 'mother', 'war', 'sister', 'parents', 'sent', 'write', \\'cross', 'found', 'red', 'brother', 'knew', 'family'\end{tabular}} \\ \hline
\multicolumn{1}{|l|}{Bar Mitzvah} & \multicolumn{1}{l|}{\begin{tabular}[c]{@{}l@{}}'bar', 'mitzvah', 'remember', 'torah', 'synagogue', 'mitzvahed', 'school', \\'jewish', 'rabbi', 'hebrew', 'shul', 'family', 'mitzvahs', 'father', '13'\end{tabular}} \\ \hline
\end{tabular}

\clearpage
\vspace{.2cm}
\centering
\begin{tabular}{ll}
\multicolumn{1}{c}{\textbf{Topic Title}} & \multicolumn{1}{c}{\textbf{Top-15 words in topic}} \\ \hline
\multicolumn{1}{|l|}{War News} & \multicolumn{1}{l|}{\begin{tabular}[c]{@{}l@{}}'radio', 'poland', 'jews', 'polish', 'war', 'news', 'jewish', 'people', 'german', 'knew', \\'germans', 'germany', 'heard', 'poles', 'warsaw'\end{tabular}} \\ \hline
\multicolumn{1}{|l|}{Greeting} & \multicolumn{1}{l|}{\begin{tabular}[c]{@{}l@{}}'thank', 'yourn', 'much', 'muchrn', 'sharing', 'welcome', 'welcomern', 'add', 'youre', \\'mrs', 'testimony', 'concludes', 'thanks', 'say', 'want'\end{tabular}} \\ \hline
\multicolumn{1}{|l|}{Legal} & \multicolumn{1}{l|}{\begin{tabular}[c]{@{}l@{}}'trial', 'trials', 'court', 'nuremberg', 'crimes', 'witnesses', 'case', 'courtroom', \\'witness', 'judge', 'cases', 'justice', 'defense', 'evidence', 'interpreter'\end{tabular}} \\ \hline
\multicolumn{1}{|l|}{Sport} & \multicolumn{1}{l|}{\begin{tabular}[c]{@{}l@{}}'play', 'soccer', 'used', 'played', 'sports', 'school', 'games', 'remember', 'sport', \\'friends', 'swimming', 'playing', 'ball', 'club', 'liked'\end{tabular}} \\ \hline
\multicolumn{1}{|l|}{Partisan} & \multicolumn{1}{l|}{\begin{tabular}[c]{@{}l@{}}'partisans', 'partisan', 'group', 'russian', 'germans', 'forest', 'killed', 'people', \\'army', 'fighting', 'woods', 'fight', 'food', 'knew', 'thats'\end{tabular}} \\ \hline
\multicolumn{1}{|l|}{Far East} & \multicolumn{1}{l|}{\begin{tabular}[c]{@{}l@{}}'shanghai', 'japanese', 'chinese', 'china', 'japan', 'hongkew', 'refugees', 'people', \\'war', 'see', 'money', 'ship', 'american', 'boat', 'harbor'\end{tabular}} \\ \hline
\multicolumn{1}{|l|}{Yellow Star} & \multicolumn{1}{l|}{\begin{tabular}[c]{@{}l@{}}'star', 'wear', 'yellow', 'wearing', 'stars', 'jewish', 'david', 'jews', 'remember', \\'armband', 'wore', 'germans', 'jew', 'people', 'think'\end{tabular}} \\ \hline
\multicolumn{1}{|l|}{Number Tattoo} & \multicolumn{1}{l|}{\begin{tabular}[c]{@{}l@{}}'number', 'tattooed', 'tattoo', 'numbers', 'auschwitz', 'camp', 'arm', 'barracks', \\'given', 'birkenau', 'tattooing', 'people', 'triangle', 'put', 'prisoners'\end{tabular}} \\ \hline
\multicolumn{1}{|l|}{Camp Selection} & \multicolumn{1}{l|}{\begin{tabular}[c]{@{}l@{}}'mengele', 'selection', 'auschwitz', 'gas', 'camp', 'right', 'people', 'saw', \\'side', 'twins', 'selected', 'see', 'left', 'told', 'experiments'\end{tabular}} \\ \hline
\multicolumn{1}{|l|}{Schindler List} & \multicolumn{1}{l|}{\begin{tabular}[c]{@{}l@{}}'schindler', 'factory', 'list', 'schindlers', 'plaszow', 'oskar', 'goeth', \\'brxfcnnlitz', 'people', 'camp', 'brunnlitz', 'inaudible', 'working', 'knew', 'auschwitz'\end{tabular}} \\ \hline
\multicolumn{1}{|l|}{Jewish Community} & \multicolumn{1}{l|}{\begin{tabular}[c]{@{}l@{}}'jewish', 'town', 'population', 'jews', 'lived', 'community', 'city', \\'people', 'synagogue', 'families', 'big', 'school', 'area', 'lot', 'business'\end{tabular}} \\ \hline
\multicolumn{1}{|l|}{Expensive Objects} & \multicolumn{1}{l|}{\begin{tabular}[c]{@{}l@{}}'jewelry', 'money', 'gold', 'ring', 'things', 'silver', 'coins', 'hide', \\'give', 'take', 'put', 'buy', 'remember', 'sold', 'whatever'\end{tabular}} \\ \hline
\multicolumn{1}{|l|}{Army} & \multicolumn{1}{l|}{\begin{tabular}[c]{@{}l@{}}'army', 'training', 'basic', 'infantry', 'drafted', 'fort', 'draft', 'corps', 'sergeant', \\'officer', 'unit', 'military', 'citizen', 'british', 'service'\end{tabular}} \\ \hline
\multicolumn{1}{|l|}{Concentration Camps} & \multicolumn{1}{l|}{\begin{tabular}[c]{@{}l@{}}'bergenbelsen', 'buchenwald', 'camp', 'people', 'belsen', 'barrack', \\'dead', 'arrived', 'saw', 'barracks', 'remember', 'block', 'liberated', 'prisoners', 'many'\end{tabular}} \\ \hline
\multicolumn{1}{|l|}{Kristallnacht} & \multicolumn{1}{l|}{\begin{tabular}[c]{@{}l@{}}'kristallnacht', 'synagogue', 'november', 'happened', '1938', 'remember', \\'father', 'arrested', 'night', 'school', 'mother', 'home', 'glass', 'jewish', 'day'\end{tabular}} \\ \hline
\multicolumn{1}{|l|}{Italy} & \multicolumn{1}{l|}{\begin{tabular}[c]{@{}l@{}}alian', 'italy', 'italians', 'athens', 'switzerland', 'rome', 'mussolini', 'people', \\'camp', 'modena', 'train', 'germans', 'bari', 'chichibo', 'nonantola'\end{tabular}} \\ \hline
\multicolumn{1}{|l|}{Birth} & \multicolumn{1}{l|}{\begin{tabular}[c]{@{}l@{}}'baby', 'hospital', 'doctor', 'child', 'pregnant', 'husband', 'mother', 'father', \\'sick', 'born', 'cancer', 'died', 'told', 'home', 'never'\end{tabular}} \\ \hline
\multicolumn{1}{|l|}{Parents} & \multicolumn{1}{l|}{\begin{tabular}[c]{@{}l@{}}'father', 'mother', 'come', 'says', 'take', 'little', 'told', 'saw', 'knew', 'see', \\'place', 'away', 'want', 'ill', 'thought'\end{tabular}} \\ \hline
\multicolumn{1}{|l|}{Scandinavia} & \multicolumn{1}{l|}{\begin{tabular}[c]{@{}l@{}}'sweden', 'denmark', 'danish', 'swedish', 'danes', 'copenhagen', 'stockholm', \\'king', 'people', 'malmo', 'jews', 'goteborg', 'jewish', 'swedes', 'government'\end{tabular}} \\ \hline
\multicolumn{1}{|l|}{University} & \multicolumn{1}{l|}{\begin{tabular}[c]{@{}l@{}}'university', 'college', 'school', 'degree', 'years', 'new', 'high', 'york', \\'job', 'engineering', 'masters', 'worked', 'work', 'social', 'graduate'\end{tabular}} \\ \hline
\multicolumn{1}{|l|}{Australia} & \multicolumn{1}{l|}{\begin{tabular}[c]{@{}l@{}}'australia', 'melbourne', 'australian', 'sydney', 'boat', 'fremantle', 'ship', \\'arrived', 'come', 'australians', 'people', 'job', 'english', 'friends', 'years'\end{tabular}} \\ \hline
\multicolumn{1}{|l|}{Holland} & \multicolumn{1}{l|}{\begin{tabular}[c]{@{}l@{}}'dutch', 'holland', 'amsterdam', 'westerbork', 'jews', 'people', 'rotterdam', \\'german', 'camp', 'germans', 'war', 'jewish', 'happened', 'nazis', 'germany'\end{tabular}} \\ \hline
\multicolumn{1}{|l|}{Christianity} & \multicolumn{1}{l|}{\begin{tabular}[c]{@{}l@{}}'catholic', 'church', 'priest', 'baptized', 'communion', 'religion', 'jewish', \\'catholicism', 'mother', 'school', 'convert', 'never', 'convent', 'prayers', 'think'\end{tabular}} \\ \hline
\multicolumn{1}{|l|}{Deportations} & \multicolumn{1}{l|}{\begin{tabular}[c]{@{}l@{}}'deported', 'deportation', 'people', 'deportations', 'jews', 'day', \\'work', 'ghetto', 'happened', 'heard', 'think', 'told', 'believe', 'sent', 'knew'\end{tabular}} \\ \hline
\multicolumn{1}{|l|}{Clothes} & \multicolumn{1}{l|}{\begin{tabular}[c]{@{}l@{}}'shoes', 'clothes', 'shower', 'hair', 'shaved', 'camp', 'clothing', 'naked', \\'put', 'auschwitz', 'pair', 'showers', 'barracks', 'barrack', 'women'\end{tabular}} \\ \hline
\multicolumn{1}{|l|}{Packing} & \multicolumn{1}{l|}{\begin{tabular}[c]{@{}l@{}}'take', 'clothing', 'little', 'suitcase', 'remember', 'things', 'knitting', \\'left', 'made', 'used', 'yarn', 'everything', 'mother', 'maybe', 'something'\end{tabular}} \\ \hline
\multicolumn{1}{|l|}{Judenrat} & \multicolumn{1}{l|}{\begin{tabular}[c]{@{}l@{}}'judenrat', 'judenrate', 'ghetto', 'police', 'jewish', 'people', 'jews', \\'germans', 'council', 'orders', 'killed', 'away', 'gestapo', 'work', 'town'\end{tabular}} \\ \hline
\multicolumn{1}{|l|}{Gypsy} & \multicolumn{1}{l|}{\begin{tabular}[c]{@{}l@{}}'gypsies', 'gypsy', 'camp', 'block', 'people', 'auschwitz', 'mengele', 'lager', \\'saw', 'jews', 'gassed', 'barracks', 'eichman', 'see', 'told'\end{tabular}} \\ \hline
\end{tabular}

\clearpage
\vspace{.2cm}
\centering
\begin{tabular}{ll}
\multicolumn{1}{c}{\textbf{Topic Title}} & \multicolumn{1}{c}{\textbf{Top-15 words in topic}} \\ \hline
\multicolumn{1}{|l|}{Kapo} & \multicolumn{1}{l|}{\begin{tabular}[c]{@{}l@{}}'kapos', 'kapo', 'prisoners', 'camp', 'block', 'barrack', 'people', 'german', 'political', 'killed', \\'work', 'put', 'prisoner', 'somebody', 'remember'\end{tabular}} \\ \hline
\multicolumn{1}{|l|}{Poland} & \multicolumn{1}{l|}{\begin{tabular}[c]{@{}l@{}}'warsaw', 'polish', 'lublin', 'place', 'train', 'find', 'army', 'walked', 'also', 'anyway', 'station', \\'says', 'poles', 'people', 'praga'\end{tabular}} \\ \hline
\multicolumn{1}{|l|}{Camp Barracks} & \multicolumn{1}{l|}{\begin{tabular}[c]{@{}l@{}}'barrack', 'barracks', 'beds', 'bunk', 'people', 'bunks', 'slept', 'camp', 'wooden', 'straw', \\'bed', 'sleeping', 'cold', 'sleep', 'little'\end{tabular}} \\ \hline
\multicolumn{1}{|l|}{Death} & \multicolumn{1}{l|}{\begin{tabular}[c]{@{}l@{}}'cemetery', 'buried', 'grave', 'jewish', 'people', 'put', 'died', 'stone', 'family', 'find', \\'mass', 'tombstones', 'graves', 'place', 'stones'\end{tabular}} \\ \hline
\multicolumn{1}{|l|}{Unknown} & \multicolumn{1}{l|}{\begin{tabular}[c]{@{}l@{}}'people', 'remember', 'jewish', 'see', 'think', 'father', 'come', 'knew', 'little', 'german', \\'camp', 'day', 'mother', 'told', 'thats\end{tabular}} \\ \hline
\end{tabular}

\end{document}